\documentclass{article}

\usepackage{PRIMEarxiv}

\usepackage[utf8]{inputenc} 
\usepackage[T1]{fontenc}    
\usepackage{hyperref}       
\usepackage{url}            
\usepackage{booktabs}       
\usepackage{amsfonts}       
\usepackage{microtype}      
\usepackage{lipsum}
\usepackage{fancyhdr}       
\usepackage{graphicx}       
\usepackage{authblk}
\graphicspath{{media/}}

\usepackage{amsmath}
\usepackage[ruled,vlined,linesnumbered]{algorithm2e}
\DontPrintSemicolon
\SetKwInput{Input}{Input}
\SetKwInput{Output}{Output}

\title{Encoding large information structures in linear algebra and statistical models}

\author{
David Banh  [1], Alan Huang [2]\\
[1] \ \ AskExplain\\
[2] \ \ The University of Queensland\\
Corresponding email: david.b@askexplain.com
}

\begin{document}
\maketitle

\begin{abstract}

Large information sizes in samples and features can be encoded to speed up the learning of statistical models based on linear algebra and remove unwanted signals. Encoding information can reduce both sample and feature dimension to a smaller representational set. Here two examples are shown on linear mixed models and mixture models speeding up the run time for parameter estimation by a factor defined by the user's choice on dimension reduction (can be linear, quadratic or beyond based on dimension specification).

GitHub at:  
\nolinkurl{https://github.com/AskExplain/encoding_information}

\end{abstract}

\section{Introduction}\label{}

Large sizes in the samples and features in statistics creates large matrix objects in linear algebra. Large information structures generally occur when the dimensional size is large enough such that the (generalised) inverse of the cross product of two matrices is computationally infeasible to calculate quickly (this generally occurs for sizes greater than n = 10,000). 

Here, an encoding of the information is proposed to reduce the matrix dimensions to a tractable size such that after inverting and decoding, gives a representational structure similar to when the operation runs on the full matrix. Examples are used for mixture models \cite{Viroli2019} and linear mixed models \cite{Yang2011}.

Encoding information can be done with Singular Value Decomposition (SVD), or, a similar method based on SVD by same authors called Generative Encoding \cite{Banh2021.07.09.451779} \cite{DBLP:journals/corr/abs-2201-08233}. 

\section{Methods}\label{}

\subsection{Linear Mixed Model}
For a linear mixed model to be encoded, first consider the mixed model equation:

$$ Y = X \beta + Z u + e $$

To encode sample information, a function is introduced as a parameter $\alpha$ to re-weight the samples into a reduced dimension:

$$ \alpha Y = \alpha X \beta + \alpha Z u + e $$

Provided each of $Y$, $X$ and $Z$ are of $n$ samples, and $\alpha$ is a parameter that transforms the $n$ samples into $m$ samples (where $n > m$), then the final model will be learned via the covariance of a smaller size.

\subsubsection{Genetic Relatedness Matrix}\label{}

For example, the general model in genetics to measure the heritability of a trait is given as:

$$ Y \sim N ( X \beta , G \sigma_g + D_1 \sigma_e ) $$

Reducing the sample size would introduce to this model, the following:

$$ \alpha Y \sim N ( \alpha X \beta , ( \alpha G \alpha^T ) \sigma_g + D_2 \sigma_e ) $$

Notice that $\alpha$ is of $m$ dimensions rather than $n$, enabling the linear mixed model to learn with a smaller Genetic Relatedness Matrix, yet still retaining the ability to learn $\sigma_g$ the heritability of the phenotype.

\subsection{Mixture Model}
Rather than encoding sample information, the dimensions of the feature structure can be encoded. For example for a mixture model to be encoded, first consider the expression:

$$ Y \sim \pi_{i} N( \mu_{i}, \Sigma_{i} ) $$

To encode sample structure, a function is introduced as a parameter $\alpha$ to re-weight the samples into a reduced dimension:

$$ Y \beta \sim \pi_{i} N(   \mu_{i} \beta  ,  \beta^T \Sigma_{i} \beta  ) $$

By reducing the dimensions of the mixture model to learn a model from $r$ features rather than $p$ features (where $p > r$) the model can be learned faster as the feature information $p$ is learned in the reduced dimensions $r$.

\subsubsection{Factor analytic models}\label{}

For example, the general model for mixtures of factor analyses in psychology or economic studies used to measure the scores of particular behaviours or events is given as:

$$ X \sim \pi_{i} N ( \mu_{i} + \Lambda_{i} z , D_{i} ) $$
$$ X|z \sim \pi_{i} N ( \mu_{i} ,  \Lambda_{i} \Lambda_{i} ^T + D_{i} ) $$

where the conditional expression based on the latent features $z$ is found in the second expression above.

Taking into account the feature encoding to reduce the dimensional size of $p$ takes the dimensions of the features to $r$ dimensions

$$ X \beta \sim \pi_{i} N ( \mu_{i} \beta + \beta \Lambda_{i} z_{\beta} , D_{i_{\beta}} ) $$

$$ X\beta |z_{\beta} \sim \pi_{i} N ( \mu_{i} \beta  ,  ( \beta \Lambda_{i} \Lambda_{i}^T \beta^T ) + D_{i_{\beta}} ) $$

\newpage

\section{Results}

\subsection{Linear Mixed Model}
To test the results on a linear mixed model, 1000 permutations were run on a simulated dataset with a heritability of 0.5.

The package GMMAT in R was used to run the analysis with a Genetic Relatedness Matrix. The table below shows the results comparing the encoded and original linear mixed models comparing the heritability estimates and runtime of the full mixed model with the encoded model (including the time to learn the encoding). 

This is for 100 permutation runs of a simulation with 1000 samples and 100 SNPs simulated according to

$$ Z \sim Binom ( m = 2 , p = 0.5 ) $$
$$ \lambda_s \sim N ( 0 , \sqrt{( h_2 / p )} ) $$
$$ Y \sim N ( Z \lambda_s , \sqrt{( 1 - h_2 )} ) $$

Where the Genetic Relatedness Matrix (1000 by 1000 dimensions) is given by:
$$ GRM = ( Z Z^T ) / p $$

\begin{figure}[h!]
    \centering
    \includegraphics[width=\textwidth]{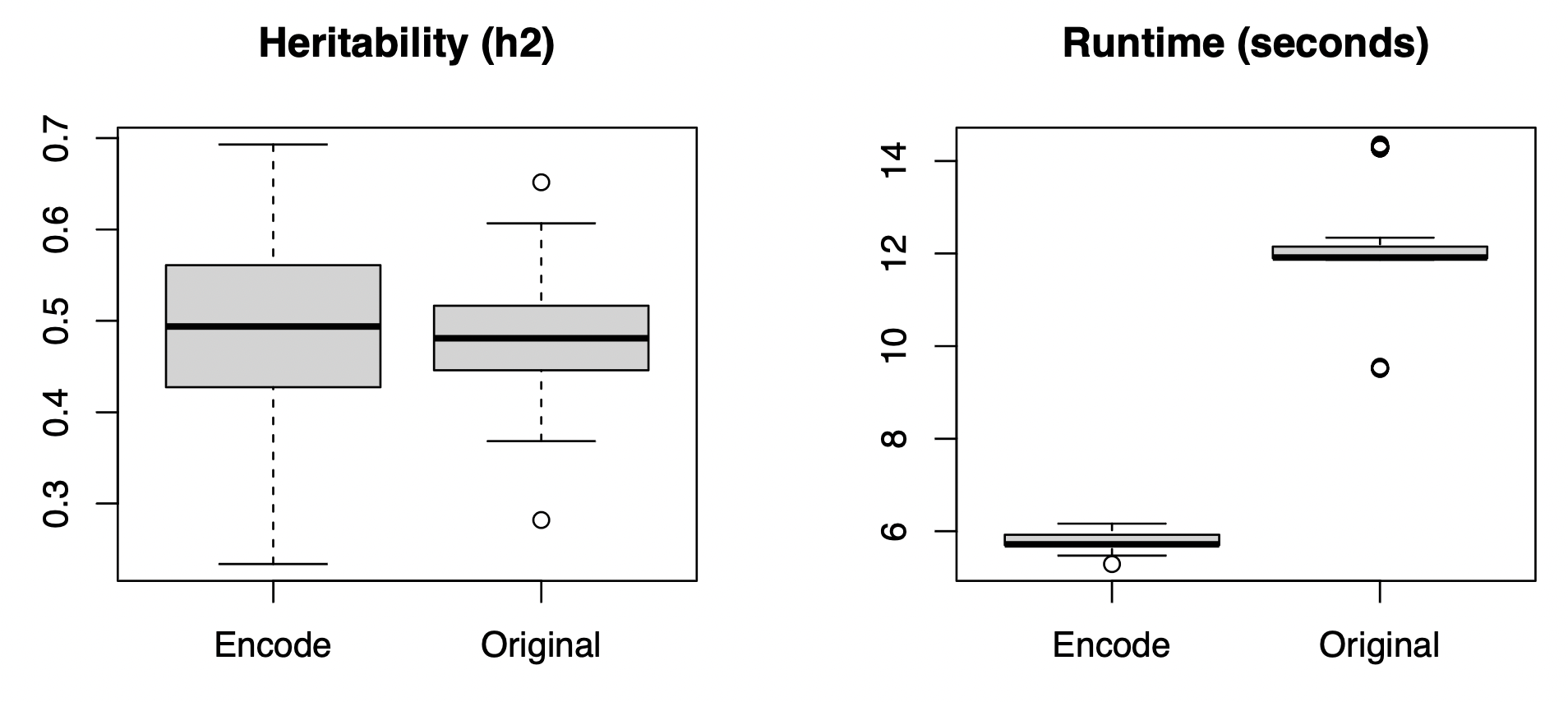}
    \caption{ Comparing an encoded and standard linear mixed model for a heritability estimate distributed around 0.5 (true is at $h^2 = 0.5$), and runtime in seconds}
    \label{fig:my_label}
\end{figure}

\subsection{Mixture Model}
To test the results on a linear mixed model, approximately 1000 permutations were run on a test dataset from the pdfCluster \cite{JSSv057i11} package using the OliveOil dataset only on the numerical dataset.

Given the OliveOil dataset has two categorical features structured hierarchically, one as a subset of the other, the categorical feature with the fewest number of categories was used. This equates to 3 known clusters to label.

The features were encoded ranging from 2 to 8 (the original number of numerical features in the OliveOil dataset). 

\begin{figure}[h!]
    \centering
    \includegraphics[width=\textwidth]{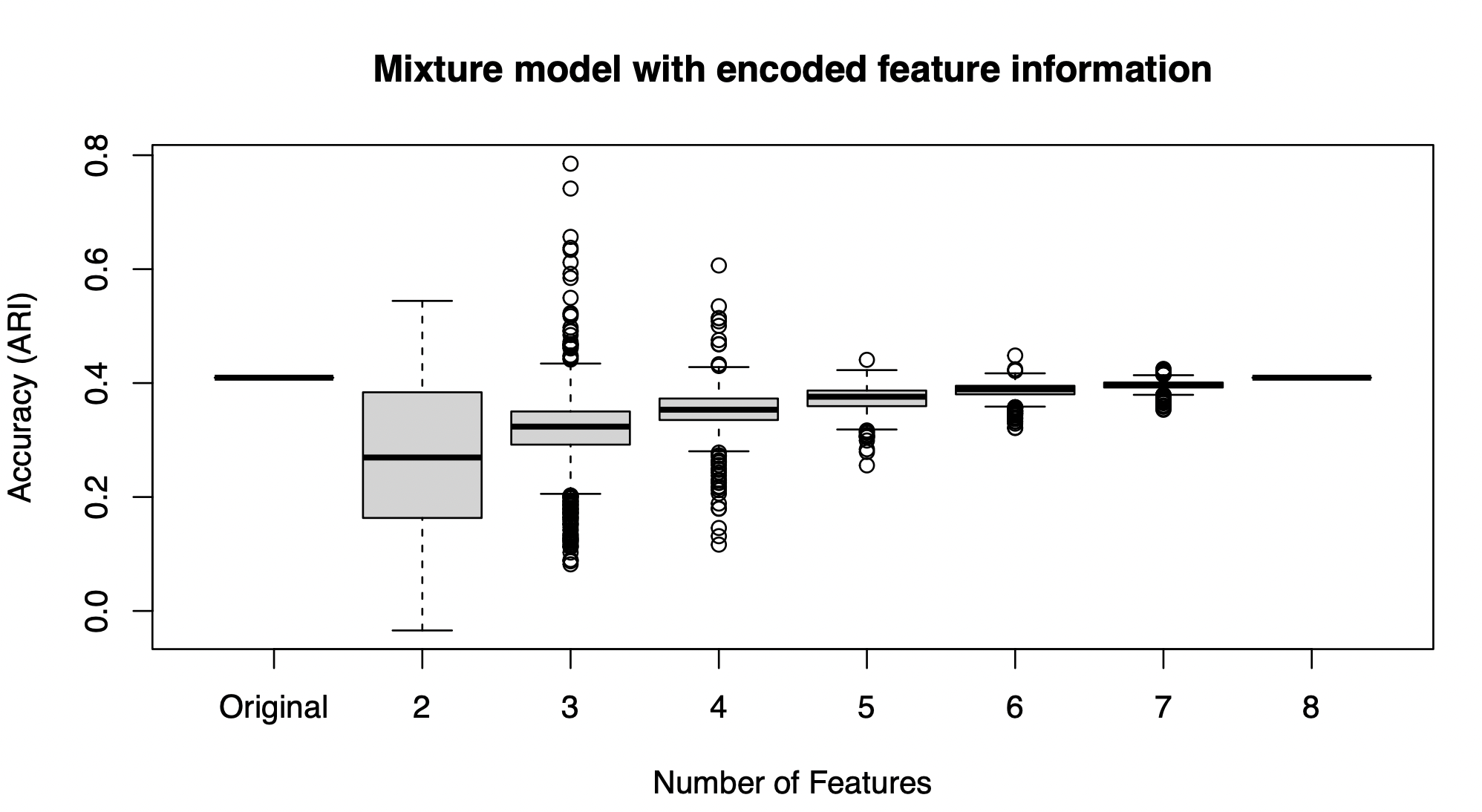}
    \caption{ Mixture model with an encoding to 2 to 8 features, compared with the original features (p = 8). }
    \label{fig:my_label}
\end{figure}

\newpage

\section{Discussion}

Given information is being encoded, it is expected for there to be information loss leading to higher variability compared to the standard linear mixed model. However, due to reduced sample size via an encoding the runtime is faster - almost half the speed of the original mixed model according to the R package GMMAT \cite{Chen2016}.

Notice that with an encoding of the features into a lower dimensional space - mixture model clustering accuracy has more flexibility due to an increase the degrees of freedom leading to fit worse or better models. On average, the larger the encoding - up to the original number of features, the higher the accuracy of the final clustering.

\section{Conclusion}

Information in a vector, matrix, or tensor object can be encoded by learning representational features that simultaneously: encodes the object and represents the object via factorisation. Through considering functions that encodes and re-represents the information via factorisation, an optimal model is learned that extracts relevant signals from the data object to manipulate the feature and, or sample structure.

This has implications on linear algebra and statistical methods - encoding the samples can reduce the computational run time for mixed models when a sample covariance matrix is used. Alternatively, features can be encoded to reduce the computational run time of a feature covariance in mixture models.

\section{Acknowledgements}

Professor Geoff McLachlan has been a tremendous help in the guidance of past work on mixture models (see Deep Gaussian Mixture Models). Professor Jian Yang has also been an inspiration for the mixed model work. Also a thanks to Yuna Zhang for jump starting the work by providing preliminary scripts on Average Information with mixed models.

\bibliographystyle{unsrt}  
\bibliography{main}

\begin{thebibliography}{1}

\bibitem{Viroli2019}
Cinzia Viroli and Geoffrey~J. McLachlan.
\newblock Deep gaussian mixture models.
\newblock {\em Statistics and Computing}, 29(1):43--51, Jan 2019.

\bibitem{Yang2011}
Jian Yang, S.~Hong Lee, Michael~E. Goddard, and Peter~M. Visscher.
\newblock Gcta: a tool for genome-wide complex trait analysis.
\newblock {\em American journal of human genetics}, 88(1):76--82, Jan 2011.
\newblock 21167468[pmid].

\bibitem{Banh2021.07.09.451779}
David Banh and Alan Huang.
\newblock Scalable parametric encoding of multiple modalities.
\newblock {\em bioRxiv}, 2022.

\bibitem{DBLP:journals/corr/abs-2201-08233}
David Banh.
\newblock Sample summary with generative encoding.
\newblock {\em CoRR}, abs/2201.08233, 2022.

\bibitem{JSSv057i11}
Adelchi Azzalini and Giovanna Menardi.
\newblock Clustering via nonparametric density estimation: The r package
  pdfcluster.
\newblock {\em Journal of Statistical Software}, 57(11):1–26, 2014.

\bibitem{Chen2016}
Han Chen, Chaolong Wang, Matthew~P. Conomos, Adrienne~M. Stilp, Zilin Li, Tamar
  Sofer, Adam~A. Szpiro, Wei Chen, John~M. Brehm, Juan~C. Celed{\'o}n, Susan
  Redline, George~J. Papanicolaou, Timothy~A. Thornton, Cathy~C. Laurie,
  Kenneth Rice, and Xihong Lin.
\newblock Control for population structure and relatedness for binary traits in
  genetic association studies via logistic mixed models.
\newblock {\em American journal of human genetics}, 98(4):653--666, Apr 2016.
\newblock 27018471[pmid].

\end{thebibliography}

\end{document}